# Translating Math Formula Images to LaTeX Sequences Using Deep Neural Networks with Sequence-level Training

Zelun Wang
*Department of Computer Science and Engineering*
*Texas A&M University*
College Station, USA
zelun@tamu.edu

Jyh-Charn Liu
*Department of Computer Science and Engineering*
*Texas A&M University*
College Station, USA
liu@cse.tamu.edu

*Abstract*—In this paper we propose a deep neural network model with an encoder-decoder architecture that translates images of math formulas into their LaTeX markup sequences. The encoder is a convolutional neural network (CNN) that transforms images into a group of feature maps. To better capture the spatial relationships of math symbols, the feature maps are augmented with 2D positional encoding before being unfolded into a vector. The decoder is a stacked bidirectional long short-term memory (LSTM) model integrated with the soft attention mechanism, which works as a language model to translate the encoder output into a sequence of LaTeX tokens. The neural network is trained in two steps. The first step is token-level training using the Maximum-Likelihood Estimation (MLE) as the objective function. At completion of the token-level training, the sequence-level training objective function is employed to optimize the overall model based on the policy gradient algorithm from reinforcement learning. Our design also overcomes the exposure bias problem by closing the feedback loop in the decoder during sequence-level training, i.e., feeding in the predicted token instead of the ground truth token at every time step. The model is trained and evaluated on the IM2LATEX-100K dataset and shows state-of-the-art performance on both sequence-based and image-based evaluation metrics.

*Keywords—deep learning; encoder-decoder; seq2seq model; image to LaTeX; reinforcement learning; math formulas*

## I. INTRODUCTION

Math formulas often carry the most significant technical substances in many science, technology, engineering and math (STEM) fields. Being able to extract the math formulas from digital documents and translate them into markup languages is very useful for a wide range of information retrieval tasks. Portable Document Format (PDF) is the *de facto* standard publication format, which makes document distribution very easy and reliable. Although math formulas can be recognized by human readers relatively easily, computer-based math formula recognition in PDF documents remains a major challenge. This is mainly because the PDF format does not contain tagged information about its math contents. Recognizing math formulas from PDF documents is intrinsically difficult because of the presence of unusual math symbols and complex layout structures. In addition, math formulas in PDF documents could partially be represented by blocks of graphics directly rendered from the PDF glyphs, which preserves the correct shapes but misses the meaning of contents. These problems would be readily solved if the markup sources of the PDF documents are available. A good example is the preprint repositories *arXiv.org* which gives readers access to the LaTeX source files along with the PDF files, but it only comprises a small fraction of the existing digital publications. For vast majority of digital documents, advanced techniques are needed to translate the PDF math contents into their markup sources. Being a structured math description language, LaTeX can be used to retrieve math formulas, and can be easily converted to other formats such as MathML [5] to support high-level applications.

With the earliest effort dating back to 1967 [6], different approaches have been developed to recover math contents with different levels of success. Recent advancement in optical character recognition (OCR) techniques has made it possible to recognize text in digital documents at high accuracy. However, recognizing math formulas is difficult, because on top of recognition of individual math symbols, it is also necessary to recognize the structural relationship among symbols, such as sub/sup-scripts, nested fractions, matrix, etc. Researchers have developed rule-based structural analysis methods and syntactic parsers to convert math formulas to their markup languages. One successful example is the INFTY system [1], which was designed to convert documents into structured formats like LaTeX, and was later made into a commercial software called InftyReader for digital document processing. With the rise of deep learning technology, it has been demonstrated that hand-crafted features and rules can now be replaced by learnable feature representations.

Translating math formula images to LaTeX sequences is a joint field of image processing and text processing, which has recently gained increased research interest in the deep learning community [7-9]. The sequence-to-sequence model (seq2seq), also called the encoder-decoder architecture, has been successfully applied to intersect these two fields. The encoder for such applications is usually a convolutional neural network (CNN) which encodes the input images as abstract feature representations, and the decoder is usually a recurrent neural network (RNN) that represents a language model to translate the encoder output into a sequence of *tokens* drawn from a vocabulary. This architecture makes the size of input images and output sequences flexible, and could be trained in an *end-to-end* fashion. Seq2seq model has been successfully used in image captioning [8, 9] and scene text recognition [7] tasks, which shares similar technical requirements with that of the image to LaTeX task. Recently, the authors in [2] successfully applied an attention-based seq2seq model to translate images to LaTeX, which demonstrated the model's capability of handling structural contents.

Leveraging the previous success, in this paper we propose a new seq2seq model called MI2LS (Math Image to LaTeX Sequence) which focuses on addressing three key problems that have not been investigated in prior works. Firstly, to help the model better differentiate the 2-dimensional spatial relationship of math symbols, we propose to augment the image feature maps by adding sinusoidal positional encoding for richer representation of spatial locality information. Secondly, we propose a sequence-level objective function based on the BLEU (bilingual evaluation understudy) [10] score, which could better capture the interrelationship among different tokens in a LaTeX sequence than the token-level cross-entropy loss. Knowing that the sequence-level evaluation score is discrete and non-differentiable, we propose to solve the optimization problem based on the policy gradient algorithm [11] in reinforcement learning for model training. Thirdly, we eliminate the exposure bias [12] problem by closing the feedback loop during the sequence-level training, i.e., feeding back the predicted token instead of the ground truth token for the next time step. This is made possible because the token alignment problem in token-level training no longer exists in sequence-level training. The overall model architecture includes a CNN encoder, an RNN decoder, and a soft attention mechanism, as shown in Figure 1. The model was trained and evaluated on the IM2LATEX-100K dataset [2], and achieved state-of-the-art performance on both the BLEU score and image similarity measurements.

## II. RELATED WORK

Automatic recognition of math formulas in digital publications has long been recognized as a challenging task [13]. The task first requires to locate math formulas in digital documents, then analyze the structure of math formulas, and finally translate them into math markup languages. In [14], Garain *et al.* proposed to use a commercial OCR tool as a text classifier, where patterns that cannot be recognized by the OCR were further analyzed to detect math formulas. In [15], *Wang et al.* developed a PDF parser to detect math formulas based on the font statistics with a feed-forward algorithm. In [16], they further proposed a bigram label regularization method to reduce the over-segmentation problem during formula detections. In [17], Gao *et al.* proposed to combine the PDF font information with vision features, and manually labeled a large dataset to train a deep neural network for math formula detection. Once math formulas are detected, the next step is to analyze their 2-dimensional layout structure. Twaaliyondo *el al.* in [18] proposed a method that first divided the formulas into subexpressions based on larger symbols and blank spaces in a recursive manner, and then represented the structure of the formulas as a tree. In [1], Suziki *et al.* used a similar approach as [14] to first locate the math formulas, and then represented the structure of math formulas as trees, and used a minimum-cost spanning-tree algorithm for the structure analysis. This proposed work was made into the commercial software -- InftyReader.

Recently, convolutional neural networks have achieved new performance levels for OCR tasks [19], which gives new solutions to translate math formulas from images in a data-driven manner, yet requiring to resolve the following additional problems: 1) the input image is not segmented, 2) the output is a sequence of tokens of arbitrary length, and 3) structural information needs to be understood. Techniques such as Connectionist Temporal Classification (CTC) [20] models the inter-label dependencies implicitly, making it possible to train a neural network directly with unsegmented data. Existing solutions to predict sequence from image inputs can be found in text recognition and image captioning tasks [7-9, 20-22], which usually combines CNN with a sequential model to construct an encoder-decoder (seq2seq) architecture. Jaderberg *et al.* in [7] showed that combining CNN with NLP techniques like Conditional Random Field (CRF) was very effective in recognizing text in images. Another common approach is to use RNN as the sequence predictor. This was referred to as a CRNN model in [22], which was end-to-end trainable for image-based sequence recognition tasks. The attention mechanism [23] has been proposed to emulate the human vision system, which allows the model to attend the salient parts of an image while generating the target sequence. Xu *et al.* in [9] combined the attention mechanism with the CRNN model which achieved further performance gain in image captioning task. With minor modifications, this architecture can be tailored to translate images of math formulas into their LaTeX markup sequences.

In [24], Zhang *et al.* proposed a gated recurrent unit (GRU) based encoder-decoder model combined with attention mechanism to translate handwritten math to LaTeX. The model takes the stroke information as inputs, and shows capability to recognize both symbols and their structures simultaneously. In [25], they replaced the GRU encoder with a CNN encoder, enabling the model to take images as inputs instead of strokes. In [2], Deng *et al.* proposed another seq2seq model that targets on machine-rendered real-world math formula images. The model is composed of a CNN and a multi-row RNN as the encoder, and an attention-based LSTM as the decoder. The model was tested on the IM2LATEX-100K dataset and outperformed the INFTY system [1]. The model was found to achieve good performance for recognizing handwritten math formulas as well [26]. In [4], Wang *et al.* improved the model in [2] by replacing the CNN encoder with a DenseNet [27], and enhanced the attention mechanism with a joint attention mechanism [28], which combines the channel-wise attention with spatial-wise attention. In [3], Zhang *et al.* increased the source image size by two times and applied double-attention mechanism, and improved the performance over [2]. All the above-mentioned works used the token-level maximum likelihood estimation as the training objective.

## III. NEURAL NETWORK ARCHITECTURE

In this section, we first present the formulation of the problem. Next, we introduce the proposed seq2seq architecture as shown in Figure 1, and explain the encoder, which is a convolutional neural network augmented with positional encoding, and the decoder, which is a stacked bidirectional long short-term memory (LSTM). In the end, we explain the soft attention mechanism.

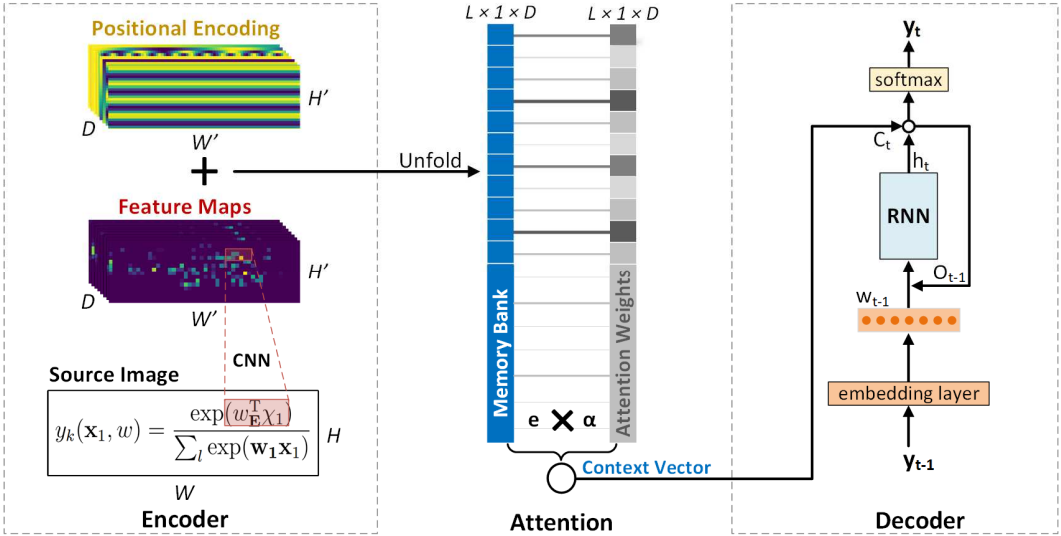

Figure 1. The proposed encoder-decoder architecture of the deep neural network.

*A. Problem Formulation*

The math formula recognition problem is formulated as a sequence prediction problem. Let $(x, y)$ be an image-LaTeX sequence pair. $x \in \mathbb{R}^{H \times W}$ is a grayscale image with height $H$ and width $W$. $y = [y_1, y_2, ..., y_T]$ is the ground truth LaTeX sequence consisting of $T$ tokens that marks up the math formula in the image. $x$ can be rendered by $y$ using the standard TeX compiler. The goal of our task is to recover $y$ given the input image $x$, i.e., to find a mapping function $f$ so that $f(x) \to y$. Given a set of $N$ image-LaTeX ground truth pairs $G = \{x^i, y^i\}_{i=1}^N$, we use supervised training to build a sequence prediction function $\hat{f}$ that approximates $f$. During the test time, we use $\hat{f}(x) \to \hat{y}$ to predict a LaTeX sequence $\hat{y}$ that reconstructs the input image $x$. Evaluation is done by measuring the similarity between $\hat{y}$ and the ground truth sequence $y$, and between the rendered image $\hat{x}$ and the ground truth image $x$.

*B. Encoder*

The encoder is used to encode the input images into abstract feature representations. It is composed of a convolutional neural network (CNN) and positional encoding.

*1) Convolutional Neural Network*

We use a CNN to extract features from the input images. CNN is consisted of convolution, pooling and activation layers. At each convolution layer, an input image is convolved with a set of kernels, which act as image filters. The kernel values are trainable, which makes the image features data-driven instead of hand-crafted. The pooling layer is usually composed of a max pooling function or average pooling function, which reduces the image size and increases the size of the receptive field. The activation layer adds nonlinearity to the neural network. It is usually a Rectified Linear Units (ReLU) that replaces negative inputs with 0 and keeps the positive inputs unchanged. We use a CNN architecture based on the VGG-VeryDeep that has been adapted particularly for OCR applications [22]. Details of the CNN configuration can be found in TABLE I. The feature maps are convolved to a 2D matrix instead of a flattened feature vector in order to retain the spatial locality information, as shown in Figure 1. This practice also allows the model to accept input images of arbitrary size. As a result of the CNN configuration, both the width $W'$ and height $H'$ of the output feature maps are 8 times smaller than that of the input image, and each position is $D$ dimensions deep ($D = 512$ in our implementation).

*2) Positional Encoding*

For text recognition, one could simply unfold the feature maps from the encoder to an array and feed it into an RNN decoder without explicitly considering spatial localization, because RNN is capable of capturing left-to-right location ordering. However, in math formulas, the spatial relationship among symbols span along different directions: left-right, top-down, sub/sup-scripts, nested, etc. The positional relationships among math symbols carry critical math semantics. As such, special efforts to preserve spatial locality are necessary. Here we tailor the 1-D positional encoding technique proposed in the Transformer model [29] to 2-D as follows:

$$PE(x, y, 2i) = \sin(x/10000^{4i/D})$$
$$PE(x, y, 2i + 1) = \cos(x/10000^{4i/D})$$
$$PE(x, y, 2j + D/2) = \sin(y/10000^{4j/D})$$
$$PE(x, y, 2j + 1 + D/2) = \cos(y/10000^{4j/D})$$

TABLE I. The encoder CNN configurations. *#maps*: the number of feature maps. *k*: kernel size. *p*: padding size. *s*: stride size. *BN*: batch normalization. The sizes are in order (height, width).

| Type | #maps | k | p | s |
|---|---|---|---|---|
| BN | | - | | |
| Convolution | 512 | (3,3) | (1,1) | (1,1) |
| MaxPooling | | (2,1) | (0,0) | (2,1) |
| BN | | - | | |
| Convolution | 512 | (3,3), | (1,1) | (1,1) |
| MaxPooling | | (1,2) | (0,0) | (1,2) |
| Convolution | 256 | (3,3) | (1,1) | (1,1) |
| BN | | - | | |
| Convolution | 256 | (3,3) | (1,1) | (1,1) |
| MaxPooling | | (2,2) | (0,0) | (2,2) |
| Convolution | 128 | (3,3) | (1,1) | (1,1) |
| MaxPooling | | (2,2) | (0,0) | (2,2) |
| Convolution | 64 | (3,3) | (1,1) | (1,1) |
| Input | Gray-scale image | | | |

, where $x$ and $y$ specifies the horizontal and vertical positions, and $i, j \in [0, D/4)$ specifies the dimension. These signals are added to the feature maps.

The positional encoding has the same size and dimension as the feature maps. Each dimension of the positional encoding is composed of a sinusoidal signal of a particular frequency and phase, representing either the horizontal or the vertical directions. We use a timescale ranging from 1 to 10000. The number of different timescales is equal to $D/4$, corresponding to different frequencies. For each frequency, we generate a sine/cosine signal on the horizontal/vertical direction. All these signals are concatenated to $D$ dimensions. The first half of the dimensions encodes the horizontal positions, and the second half encodes the vertical positions.

The positional encoding and the feature maps are added together, and then unfolded into a 1-dimensional array $\vec{E} \in \mathbb{R}^{L \times 1 \times D}$, where $L = H' \times W'$ is the length of the array. Each vector $e_i \in \vec{E}$ has a dimension of $D$, which is the feature size. Each such vector corresponds to a certain part of the input image. Note that this position encoding technique has the advantage of not adding new trainable parameters to the neural network. Furthermore, compared to trainable positional embedding, sinusoidal encoding can be scaled to lengths that are unseen in the training data.

## C. Decoder

RNN is well suited for sequence prediction tasks, because it maintains a history of the previous predictions and is able to traverse from the start to the end of sequence at arbitrary length. Let $\vec{V}$ be the vocabulary that contains all the permissible LaTeX tokens. We use an RNN to approximate a language model $p(y_t|y_1, \dots, y_{t-1}, \vec{E})$, which makes a prediction on the probability distribution of the token $y_t \in \vec{V}$ at time $t$ based on the prediction history $y_{i<t}$ and the encoder output $\vec{E}$. Next we introduce the token representation and the structure of the RNN.

### 1) Token embeddings

A LaTeX token refers to a processing unit within a LaTeX sequence to facilitate the design of the formula translator. Assuming that the LaTeX source is already split into tokens $y_1, y_2, \dots, y_T$. Details of LaTeX tokenization can be found in section V.A. A token can be fed into the RNN in different representations. One straightforward option is to represent each token as a *one-hot* vector, which implies that tokens are orthogonal to each other, and thus it may miss important language semantics. Similar to natural language words, many LaTeX tokens are interrelated. For example, '{' and '}' may have a higher correlation because they need to be used in pair based on the LaTeX grammar. As a result, we propose to add a *word embedding* [30] layer commonly used in NLP, where a token $y_t$ is projected into a high-dimensional vector $w_t$:

$$w_t = embedding(y_t)$$

This embedding is trainable and is able to capture the interrelationship between different tokens [30].

### 2) Stacked Bidirectional LSTM

We propose to use a decoder model based on two layers of bidirectional long-short term memory (LSTM) cells [31]. Stacking multiple layers of LSTM increase the depth of the RNN and thus helps to capture more complex language semantics. Using bidirectional cells in each layer helps to capture the contexts from both forward and backward directions between tokens. Figure 2(a) shows the structure of the stacked bidirectional LSTM that we used. For convenience, we will simply refer to this network as RNN henceforth.

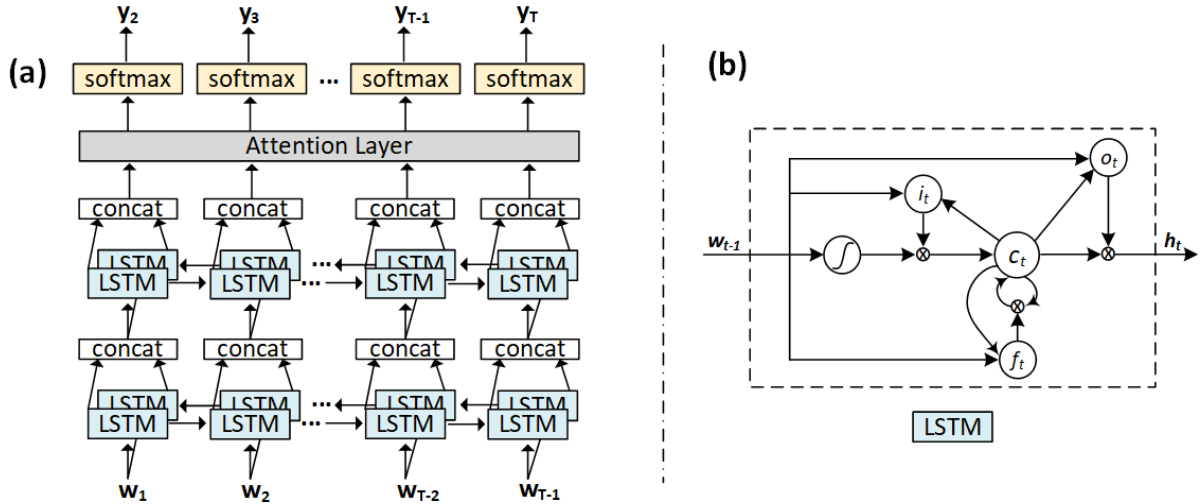

Figure 2. (a) The structure of the stacked bidirectional LSTM with attention layer. (b) The structure of an LSTM cell, where $i, f, o$ represent input gate, forget gate, and output gate separately.

LSTM is more capable of handling long sequences than the standard RNN, which is subject to the vanishing gradient problem [32] with the growth of sequence length. Figure 2(b) shows the structure of an LSTM cell. The core to the LSTM is the cell state $c_t$ that records the information that has been observed at time $t$. The LSTM is capable of adding or removing information from the cell state via three types of gates: input gate $i_t$, forget gate $f_t$, and output gate $o_t$. As implied by their names, these gates control *read* of the current input, *forget* of the current cell state value, or *output* of the current cell value. Each gate is comprised of a sigmoid neural network layer and a pointwise multiplication, expressed as below:

$$i_t = \sigma(W_{ix}w_{t-1} + W_{ih}h_{t-1})$$
$$f_t = \sigma(W_{fx}w_{t-1} + W_{fh}h_{t-1})$$
$$o_t = \sigma(W_{ox}w_{t-1} + W_{oh}h_{t-1})$$
$$c_t = f_t * c_{t-1} + i_t * \tanh(W_{cx}w_{t-1} + W_{ch}h_{t-1})$$
$$h_t = o_t * c_t$$

, where $h_t$ represents the RNN hidden state at time $t$, $\sigma$ represents the sigmoid function, and $W$ represents the weight matrix.

In NLP applications, the initial hidden state and cell state of the decoder is usually the output of the encoder RNN. However, in our model the encoder is a CNN which does not yield such an output. In order to derive informative initial states for the RNN decoder, we add additional layers to train the initial states based on the encoder output as below:

$$h_0 = \tanh(W_h \left(\frac{1}{L}\sum_{i=1}^{L} \vec{e}_i\right) + b_h)$$

$$c_0 = \tanh(W_c \left(\frac{1}{L}\sum_{i=1}^{L} \vec{e}_i\right) + b_c)$$

*D. Attention*

Theoretically, LSTM can be scaled up to capture long-term memory as needed. However, it is not uncommon that the markup of a complicated math formula extends to over a hundred LaTeX tokens. In such cases, an initial hidden state vector in RNN would be insufficient to compress all the information from the encoder. This problem is even more profound in our model because the CNN encoder does not have memory capability. The attention mechanism [23] has been introduced to solve this problem and has now become a widely adopted approach to enhance the performance on longer sequences. Basically, it maintains the complete encoder output, namely, the memory bank $\vec{E}$, based on which to calculate a context vector $C_t$ for the decoder at every time step $t$. We adopt the soft attention mechanism, which means that the context vector $C_t$ is calculated as a linear combination of the vectors $e_i \in \vec{E}$ in the memory bank:

$$C_t = \sum_{i=1}^{L} \alpha_{it} e_i$$

, where $\alpha_{it}$ is the $i^{\text{th}}$ weight at time $t$.

The attention weights are calculated with an additional feedforward layer by feeding in the previous hidden state of the LSTM $h_{t-1}$ and the memory bank $\vec{E}$, and then pass it through a softmax layer for normalization:

$$a_{it} = \beta^T \tanh(W_1 h_{t-1} + W_2 e_i)$$
$$\alpha_{it} = \text{softmax}(a_{it})$$

, where the softmax function is used to generate a probability distribution that sums up to 1, defined as $\text{softmax}(a_{it}) = \exp(a_{it}) / \sum_{k=1}^{L} \exp(a_{kt})$. The attention weights indicate which parts of the memory bank should be focused on at the current time step, thus helps the model better capture the salient parts of the input image.

To incorporate the context vector $C_t$ information into the RNN, we compute another hidden state vector $O_t$ based on the context vector $C_t$ and the current hidden state $h_t$. $O_t$ is called an attentional hidden state vector, which is fed back into the next time step of the RNN. It is also used to compute the probability distribution of the next token:

$$O_t = \tanh(W_3[h_t, C_t])$$

We also adopt the input-feeding approach proposed in [23], in which the input embedding vector is concatenated with the attentional vector from the previous time step as the input for the RNN. This way, decisions are made by considering the past alignment information.

$$h_t = RNN(h_{t-1}, [w_{t-1}, O_{t-1}])$$

The prediction probability becomes:

$$p(y_t) = \text{softmax}(W_4 O_t)$$

, which represents the probability distribution of the next token over the vocabulary $\vec{V}$.

IV. TRAINING OBJECTIVES

An ideal objective function should be constructed at the sequence level because of the rigorous nature of the LaTeX grammars. We also note that the backpropagation algorithm needs to follow the gradient direction of the objective function to update model parameters. As such, in addition to being a precise measurement of the prediction quality, it is highly desirable that the objective function is differentiable. In this section we will describe the design of a sequence-level objective function and techniques to compute its derivative based on the policy gradient algorithm. We note that it is infeasible to train the neural network with the sequence-level objective function from a random start, because the neural network may not converge under a poor random prediction policy. To overcome these challenges, we start off by training the neural network with a token-level objective function until it converges. This forms the initial state for the sequence-level training, as such the model can focus on a much smaller search space.

## A. Token-level Objective Function

The objective function of the token-level training is based on the maximum likelihood estimation (MLE). Given a training dataset of image and LaTeX sequence pairs $\{x^i, y^i\}_{i=1}^{N}$ of size $N$, where $x^i$ and $y^i$ represents the $i^{th}$ input image and ground truth LaTeX sequence respectively, the goal is to find a set of parameters $\theta$ that maximizes the log-likelihood of the training data:

$$\hat{\theta}_{MLE} = \underset{\theta}{\mathrm{argmax}}\{L_{MLE}(\theta)\}$$

, where

$$L_{MLE}(\theta) = \sum_{i=1}^{N} p(y^i, x^i)$$
$$= \sum_{i=1}^{N} \sum_{t=1}^{T} p(y_t^i | y_1^i, \dots, y_{t-1}^i, x^i)$$

This is equivalent to minimizing the cross-entropy loss (XENT):

$$L_{XENT}(\theta) = -\frac{1}{N}\left(\sum_{i=1}^{N} y^i \cdot \log(\hat{y}^i)\right)$$

, where $\hat{y}^i$ is the prediction. The derivative of the cross-entropy loss can be directly used as the gradient.

The token-level objective function faces two limitations. Firstly, it maximizes the probability of the next correct token, without considering the sequence-level contexts governed by the LaTeX grammar. Secondly, to avoid the token misalignment problem, the ground truth token needs to be fed into the RNN at every time step during the training time, instead of using the RNN's previous prediction. At the prediction time, however, the previous prediction from the RNN is fed back as the next input since the ground truth data is no longer available. As a result, the probability distribution being trained on is $p(y_t|y_{i<t}, \vec{E})$, but the probability distribution being tested on is $p(y_t|\hat{y}_{i<t}, \vec{E})$. This discrepancy is known as the *exposure bias* [12] problem. The sequence-level training objective function aims to overcome these problems.

## B. Sequence-level Objective Function

The formulation of a sequence-level training objective starts with its sequence-level performance metrics. Let $(x^i, y^i)$ be the $i^{th}$ training pair, and $\hat{y}^i$ be the prediction. Let $R(y^i, \hat{y}^i) \rightarrow [0,1]$ be a function that maps the predicted sequence to a scalar reward, where a larger value indicates a better performance. $R(y^i, \hat{y}^i)$ could be the BLEU score or any other evaluation metrics. The optimization goal is to maximize the expected reward across the dataset:

$$L_R(\theta) = \sum_{i=1}^{N} \mathbb{E}_{p_\theta(\hat{y}^i|x^i)}[R(y^i, \hat{y}^i)]$$
$$= \sum_{i=1}^{N} \sum_{\hat{y}^i \in Y(x^i)} p_\theta(\hat{y}^i|x^i) R(y^i, \hat{y}^i)$$

, where $\mathbb{E}(\cdot)$ denotes the expectance and $Y(x^i)$ is the set of all the possible predicted sequences for the input image $x^i$. The training objective becomes:

$$\hat{\theta}_R = \underset{\theta}{\mathrm{argmax}}\{L_R(\theta)\}$$

This sequence-level objective function aims to optimize the prediction of individual tokens within the context of the entire sequence. It also makes it possible to eliminate the exposure bias problem because the optimization is no longer based on each individual token but on the entire sequence, thus it is no longer necessary to feed in ground truth token at every time step to guarantee token alignment. We can simply close the feedback loop by feeding the predicted token instead of the ground truth token to the next time step during the training time.

Notice that it is computationally infeasible to optimize $L_R(\theta)$ based on exhaustive search due to the exponential growth of the search space of $Y(x^i)$. Meanwhile the gradient descent is not directly applicable here because the reward function $R(y^i, \hat{y}^i)$ is a discrete function of the prediction thus is not differentiable. To address this problem, recent solutions have been proposed in NLP community [12, 33, 34], which proposes to formulate this optimization problem as a reinforcement learning problem. In this setting, the prediction model is treated as an *agent*. Prediction on the next token is an *action*. At completion of the prediction, the predicted sequence is compared against the ground truth sequence to get a sequence-level evaluation score, which is the *reward*. The parameters of the neural network define a *policy*. Even though $L_R(\theta)$ is not differentiable, the policy gradient algorithm [11] can be used to transform the gradient of expectation as an expectation of gradient so that we can avoid taking derivative over the reward function:

$$\nabla_\theta L_R(\theta) = \sum_{i=1}^{N} \mathbb{E}_{p_\theta(\hat{y}^i|x^i)}[R(y^i, \hat{y}^i)\nabla_\theta \log p_\theta(\hat{y}^i|x^i)]$$

In principle, one may leverage the REINFORCE algorithm [35] to estimate the above expectation based on sampling methods. In specific, the expected value can be approximated by taking one sample from the distribution $\tilde{y} \sim p_\theta(y|x^i)$ using multinomial sampling [36]. Unfortunately, it difficult for the neural network to converge this way due to the high variance in gradient estimation. One technique to reduce the variance is to subtract an average reward $\bar{r}$ from the prediction reward [12]. This way, the estimated derivative becomes:

$$\tilde{\nabla}_\theta L_R(\theta) = \sum_{i=1}^{N} [R(y^i, \tilde{y}) - \bar{r}] \cdot \nabla_\theta \log p_\theta(\tilde{y}|x^i)$$

The average reward $\bar{r}$ was estimated by training a separate neural network layer in [12]. In our work, we simply use Monte Carlo sampling to estimate $\bar{r}$, i.e., taking $k$ samples from the multinomial distribution and calculate the average value. Now that the derivative is obtainable, the backpropagation algorithm can be used for the sequence-level training.

## V. EXPERIMENTS

In this section, we will first introduce the dataset used to train and evaluate our model, and then discuss the evaluation metrics and other baseline methods, followed by implementation details in the end.

### A. Dataset and Preprocessing

We used the public dataset IM2LATEX-100K [2], which is constructed from the LaTeX sources of publications crawled from *High Energy Physics - Theory* topic on arXiv.org. The dataset contains a total of 103,556 LaTeX sequences representing different math formulas. The length of characters of each sequence ranges from 38 to 997, with mean 118 and median 98. Each math formula is rendered into the PDF format by the pdfLaTeX[1] tool, and then converted to greyscale images in PNG format at resolution 1654 × 2339. The dataset is separated into a training set of 83,883 formulas, a validation set of 9,319 formulas, and a test set of 10,354 formulas.

The training of our model starts with constructing a token vocabulary $\vec{V}$. This can be done by tokenizing the LaTeX sources in the dataset. A straightforward approach to tokenize the LaTeX sources is to treat each individual character as a token. A more sophisticated approach is to parse the LaTeX sources into shortest reserved LaTeX words. For example, '\psi' stands for "$\psi$" in LaTeX, which would be treated as one single token, rather than four separate tokens '\', 'p', 's', 'i'. The second approach has the obvious advantages of reducing the sequence length and avoiding unnecessary prediction errors and computations. However, this approach is not trivial because it needs to have a complete list of LaTeX reserved words and an effective parsing algorithm to segment the LaTeX sources. Here we adopt the LaTeX parser developed in [2]. This parser first converts a LaTeX source into an abstract syntax tree using KaTex [37], and then generates the tokens by traversing through the syntax tree. One can also apply tree transformation during this process to normalize the LaTeX sequences. This normalization step can reduce the LaTeX polymorphic ambiguity since a same math formula image can be produced from different LaTeX source sequences. Details of the normalization rules can be found in [2]. Two utility tokens *<START>* and *<END>* are added to the vocabulary to represent the *start of sequence* and *end of sequence* respectively. The decoder is initialized with the <START> token and keeps making predictions until it encounters the <END> token. We end up with a vocabulary of size of 483.

Images are preprocessed by being cropped to only the formula area, and then downsampled to half of their original sizes for memory efficiency. To facilitate parallelization, images of similar sizes are grouped and padded with whitespaces into buckets of 20 different sizes.[2]

### B. Evaluation Criteria and Baselines

Two types of performance metrics are used to measure the accuracy of the prediction system. The first is the BLEU score between the predicted sequence and the ground truth sequence. Widely used to measure the quality of machine translation on natural languages, the BLEU score measures overlapping of *n*-grams. We report the cumulative 4-gram BLEU score commonly used in the literature. Due to the LaTeX grammar ambiguity, (e.g., $x_i^j$ can be expressed by either x_i^j or x^j_i), we further report the similarity between the ground truth image and the image rendered from the predicted LaTeX sequence based on three different metrics: image edit distance, exact match, and exact match without space. The image edit distance refers to the column-wise edit distance between the ground truth image and the tested image. To calculate the image edit distance, we first binarize the image, and convert the image into a 1D array. Each element in the array is a string representation of that column of data (the string is composed of 0's and 1's). The *edit distance score* is equal to $1 - e$, where $e$ is the total number of edit operations divided by the length of the 1D array. We also report the exact match accuracy (i.e., two images are exactly the same), and the exact match after eliminating the whitespace columns.

Based on these performance metrics, our method is evaluated against the commercial software InftyReader [1], and three recent works based on deep learning: WYGIWYS [2], Double Attention [3], and DenseNet [4].

### C. Implementation Detalis

Given a relatively small vocabulary size of 483, we choose a small token embedding size of 32. The dimension of the CNN feature maps and that of the RNN hidden states are both set to $D = 512$. The mini-batch gradient descent algorithm with Adam optimizer [38] is used to train the parameters, with an initial learning rate of 0.1. Batch size is set to 16 due to GPU memory limitation. To reduce overfitting and improve generalization, the dropout technique [39] with dropout rate of 0.4 is used during training. Randomly dropping out nodes during training can be viewed as a form of simulation to create an ensemble of different neural network configurations.

For sequence-level training, the choice of evaluation metrics is very flexible. For computation efficiency, we use BLEU score as the sequence-level evaluation metric. The initial learning rate is set to 0.00005 for the reinforcement training. The sampling size $k$ for calculating the average reward is set to 20.

To reduce the chance of being trapped at suboptimal solutions, beam search [40] is used while making predictions during the test time. At every time step, beam search selects $b$ tokens with the highest probabilities from the vocabulary. The model stops making new predictions until all $b$ predicted tokens become <EOS>. We use a beam size $b = 5$.

The overall system is implemented in PyTorch [41] to produce a deep learning model consisting of 10,870,595 parameters. It is trained on an 8GB NVIDIA Quadro M5000 GPU with 2048 CUDA cores.

---

[1] LaTeX (version 3.1415926-2.5-1.40.14)
[2] Different sizes of width-height buckets (in pixel): (320, 40), (360, 60), (360, 50), (200, 50), (280, 50), (240, 40), (360, 100), (500, 100), (320, 50), (280, 40), (200, 40), (400, 160), (600, 100), (400, 50), (160, 40), (800, 100), (240, 50), (120, 50), (360, 40), (500, 200).

## VI. RESULTS AND DISCUSSIONS

### A. General Performance

The detailed performance results are reported in Table II, where the last two rows show the performance of our model without and with the sequence-level reinforcement training. All the four deep learning models achieved a significantly better performance over the InftyReader system. Among different deep learning models, [3] and [4] achieved a better performance over the deep learning baseline [2], which is attributed to the introduction of more sophisticated convolutional networks and attention models. The best performance is achieved by training our model with BLEU score as the reinforcement reward, which shows the highest score on all the four evaluation metrics, with a BLEU score of 90.28%, image edit distance of 92.28%, exact match rate of 82.33%, and exact match rate without whitespace of 84.79%. The performance results reaffirm our observation about the importance of preserving positional locality, sequence-level optimization criteria, as well as the elimination of the exposure bias problem.

Next, we report a robustness analysis of our model vs. WYGIWYS [2] with respect to the sequence length. We use a bin size of 10 to quantize the sequence lengths, and report the average of the image edit distances within a bin as the performance metric. The results of the two models are shown in Figure 3. As expected, the performance of both models declines as the sequence length increases, but at significantly different rates. Knowing that the training set does not contain sequence longer than 150 tokens, this means that the models are also tested on samples with unseen lengths during the test time. At sequence length of 150, the edit distance scores of ours vs. [2] are 0.79 and 0.43, respectively, and at the length of 200, the two scores are 0.54 and 0.17 respectively. Our model shows the capability to handle sequences of unseen length better than the baseline model, especially in the range within 300. Notice that only 3.4% of the test samples have a length longer than 150 tokens, as indicated by the histogram of the token lengths shown in black curve, which makes the performance score after 150 spiky because of the data sparsity. The extra-long LaTeX sequences usually corresponds to large matrices or multi-line math formulas. It remains an open problem to translate them reliably.

In terms of computation cost, the model is first trained for 23 epochs with the MLE as the objective function, which took around 16 hours. The model with the highest token-level accuracy on the validation set is chosen as the candidate model for the sequence-level training. After we switched to the sequence-level objective function, the model is trained for another 15 epochs, which took around 75 hours. The best model

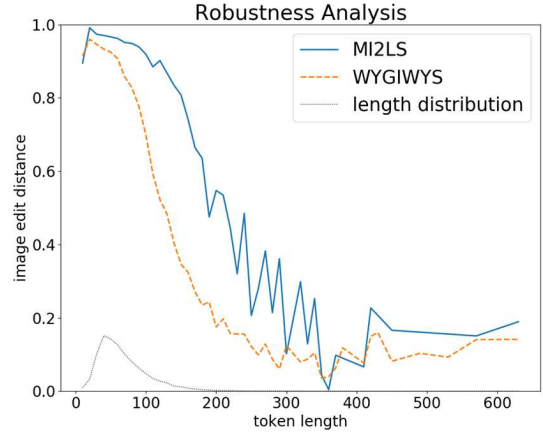

Figure 3. Robustness analysis on token length vs. image edit distance with different models. The black curve shows the density distribution of token lengths in the test set.

was selected as the one with the highest BLEU score on the validation set.

### B. Discussion

The training is end-to-end, which means no explicit information is given about segmentation of symbols in the images, scanning direction of the images, or the grammar for the LaTeX sequence outputs. And the evaluation performance suggests that these information can be learned implicitly by our deep learning model. In Figure 4, we give an example that could help us better understand the translation process of our model. The red rectangles in the images show the weights of the soft attention, while deeper color indicates higher weight values. Since the weights are applied on the CNN feature map, each attention weight corresponds to an area of 8×8 pixels in the original image, which is roughly the size of one character. We observe that the trained deep neural network can segment the symbols of different shapes and sizes, some of which are stacked or overlapped, e.g., the superscript "2" inside the square root under the fraction line. The translation process roughly follows a left-to-right order, similar to text recognition. Furthermore, it can also go from top-to-down (e.g., numerator to denominator) or down-to-top (e.g., lower to higher limits in the integral operator). This demonstrates the importance of capturing the spatial locality information. In addition, tokens that are not visible in the input images are also generated. For example, '_', '^' are generated for structural representation. '{', '}' are generated for grouping. At every time step, the weights are concentrated on only a few neighborhood regions. The model does not focus on the whitespaces until it reaches the end, in which cases an <EOS> token is generated.

TABLE II. Performance evaluation of different models on the IM2LATEX-100K dataset.

| Model | BLEU | Image Edit Distance | Exact Match | Exact Match (-ws) |
|---|---|---|---|---|
| INFTY [1] | 66.65 | 53.82 | 15.60 | 26.66 |
| WYGIWYS [2] | 87.73 | 87.60 | 77.46 | 79.88 |
| Double Attention [3] | 88.42 | 88.57 | 79.81 | - |
| DenseNet [4] | 88.25 | 91.57 | - | - |
| MI2LS w/o Reinforce | 89.08 | 91.09 | 79.39 | 82.13 |
| MI2LS with Reinforce | **90.28** | **92.28** | **82.33** | **84.79** |

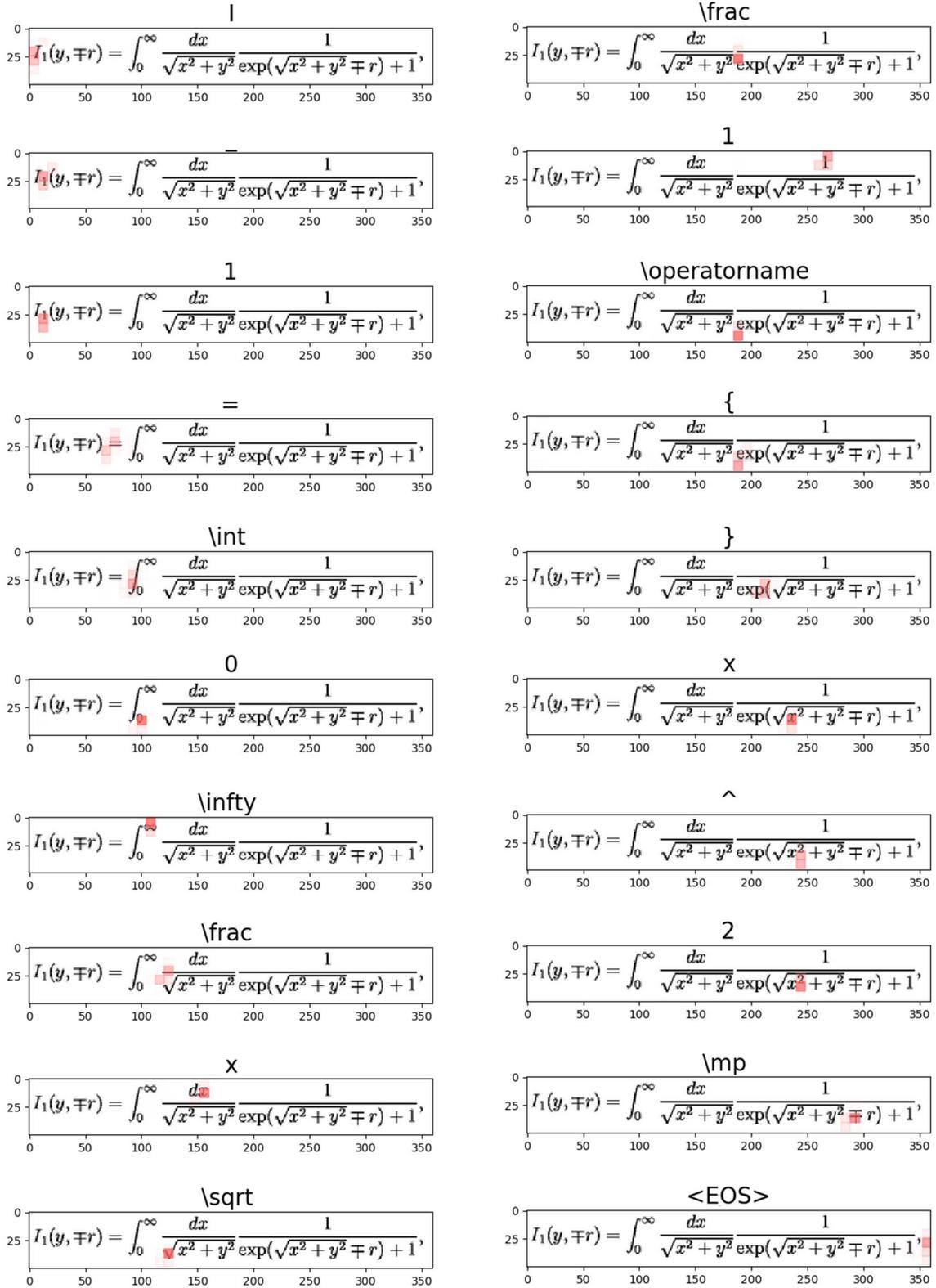

Figure 4. Visualization of the translation process for an input image. The image sequences are ordered vertically. The title of each image represents the token being produced at that certain time step. The red rectangles represent the attention weights. Darker color indicates a larger weight. We sampled 20 out of 77 predicted LaTeX tokens for concise presentation.

Notice that compared to the DenseNet model [4], our model achieved a higher performance gain on the BLEU score by 2.03%, but a lower performance gain on the image edit distance by 0.71%. A possible explanation is that the sequence-level evaluation metric we used for reinforcement learning is the BLEU score. This would naturally lead to an improvement on the BLEU score performance, but does not necessarily lead to the same amount of improvement on the image edit distance because of the polymorphic ambiguity in LaTeX language. Granted, the image edit distance score of course can be used for sequence-level training, but its drastically increased computing cost makes it an unattractive option, because every LaTeX sequence needs to be compiled to PDF and then converted from PDF to image, which requires a lot of file-level I/O, not to mention the high cost of calculating the image edit distance. One possible future improvement is to distribute this part of computation to a group of machines to facilitate reinforcement training using image edit distance. Notice that unlike [3] and [4], our performance gain over baseline [2] is attributed to adding positional encoding, introducing the sequence-level training objective, and eliminating exposure bias. We believe that our model could be potentially further improved by fusing more recent advancement in deep learning techniques, such as using DenseNet [27] as the encoder, joint attention [28] as the attention mechanism, and GRU [42] or Transformer [29] as the decoder.

## VII. CONCLUSIONS

We have proposed a deep neural network model with encoder-decoder architecture to translate images with math formulas into their LaTeX source sequences. Our model was trained in an end-to-end manner without explicit labels about image segmentation and grammar information. Nevertheless, the model managed to learn to produce LaTeX output sequence that can reproduce the input image. Using the BLEU score as the reward function and the policy gradient algorithm in reinforcement learning, we successfully trained the neural network with sequence-level objective function and eliminate the exposure bias problem. We tested the model on the IM2LATEX-100K dataset and compared it with four state-of-the-art solutions, and showed the best performance on both sequence-based and image-based measurements. The model also showed more robust performance towards longer LaTeX sequences.